\def\year{2022}\relax
\documentclass[letterpaper]{article} 
\usepackage{aaai22}  
\usepackage{times}  
\usepackage{helvet}  
\usepackage{courier}  
\usepackage[hyphens]{url}  
\usepackage{graphicx} 
\usepackage{natbib}  
\usepackage{caption} 
\DeclareCaptionStyle{ruled}{labelfont=normalfont,labelsep=colon,strut=off} 
\usepackage{algorithm}
\usepackage{algorithmic}
\usepackage{amsmath}
\usepackage{multirow}
\usepackage{amsfonts}
\usepackage{makecell}

\def\bth{\bs{\theta}}

\def\by{\bs{y}}

\def\bx{\bs{x}}

\def\bI{\bs{I}}

\newcommand{\pith}{\pi_{\bth}}

\usepackage{color}
\definecolor{gr}{gray}{0.80}
\usepackage{graphics}
\usepackage{graphicx}
\newcommand{\bs}{\boldsymbol}
\def\bx{\bs{x}}

\newcommand{\Figure}[1]{Fig.~\ref{#1}}
\newcommand{\Table}[1]{Table~\ref{#1}}
\newcommand{\Section}[1]{\S\ref{#1}}

\usepackage{newfloat}
\usepackage{listings}
\floatstyle{ruled}
\newfloat{listing}{tb}{lst}{}
\floatname{listing}{Listing}
\title{Embodied Learning for Lifelong Visual Perception}

\author{\Large \textbf{David Nilsson\textsuperscript{\rm 1}, Aleksis Pirinen\textsuperscript{\rm 2}, Erik G\"artner\textsuperscript{\rm 1}, Cristian Sminchisescu\textsuperscript{\rm 1,3}}\\} 

\affiliations{
    \textsuperscript{\rm 1}Department of Mathematics, Faculty of Engineering, Lund University\\
    \textsuperscript{\rm 2}RISE Research Institutes of Sweden\\
    \textsuperscript{\rm 3}Google Research\\
    \{david.nilsson, erik.gartner, cristian.sminchisescu\}@math.lth.se, aleksis.pirinen@ri.se
}

\begin{document}

\maketitle

\begin{abstract}
    We study lifelong visual perception in an embodied setup, where we develop new models and compare various agents that navigate in buildings and occasionally request annotations which, in turn, are used to refine their visual perception capabilities. The purpose of the agents is to recognize objects and other semantic classes in the whole building at the end of a process that combines exploration and active visual learning. As we study this task in a lifelong learning context, the agents should use knowledge gained in earlier visited environments in order to guide their exploration and active learning strategy in successively visited buildings. We use the semantic segmentation performance as a proxy for general visual perception and study this novel task for several exploration and annotation methods, ranging from frontier exploration baselines which use heuristic active learning, to a fully learnable approach. For the latter, we introduce a deep reinforcement learning (RL) based agent which jointly learns both navigation and active learning. A point goal navigation formulation, coupled with a global planner which supplies goals, is integrated into the RL model in order to provide further incentives for systematic exploration of novel scenes. By performing extensive experiments on the Matterport3D dataset, we show how the proposed agents can utilize knowledge from previously explored scenes when exploring new ones, e.g. through less granular exploration and less frequent requests for annotations. The results also suggest that a learning-based agent is able to use its prior visual knowledge more effectively than heuristic alternatives.
\end{abstract}

\section{Introduction}
Over the last decade we have witnessed rapid progress in deep learning-based visual perception \cite{KrizhevskyEtAl-nips-2012, simonyan2014very, he2016deep, ren2015faster, long2015fully}. Impressive as these developments have been on their own, many contemporary state-of-the-art perception models may nevertheless deteriorate in performance (or altogether fail) if attached e.g. to embodied systems which operate in unusual or entirely new circumstances \cite{ammirato2017dataset,pei2017towards,engstrom2019exploring}. Even when a given perception system does not break down entirely, it may still produce poorer predictions compared to the training and evaluation settings for which it was originally developed. One possible direction to limit model overfitting -- and which is often pursued -- is to work with ever larger datasets, but such a procedure is typically associated with a high cost due to the need to gather and annotate massive amounts of data. Moreover, the expressive capacity of even the deepest perception model is inherently limited, so it may be better to focus the model training on data which is most relevant for the conditions where the model is to be used.

The above motivates the need for principled and adaptive methods that can effectively refine visual perception systems in case forms of supervision become available (e.g. from a human-in-the-loop) in the present use-cases of those systems. In this work we develop and study methods for lifelong perception refinement in an embodied context, where a visual perception model trained in previously explored environments is to be refined such that it becomes as accurate as possible in a new context. A naive approach for this would be one where an embodied agent looks around and queries for annotations for all viewpoints observed in the new environment it explores. This would however require an impractically large amount of annotation and is thus clearly not feasible in practice. Moreover, this type of approach is particularly wasteful since we assume that the agent has already obtained a coarse perception model a priori. Depending on the agent's prior experience, the different viewpoints it observes in its new environment may be of highly varying importance. For example, viewpoints which contain large portions of the same semantic classes and/or appearances as the agent has seen in the past may not have to be annotated, and thus the agent may focus on those novel parts of the scene which it is most uncertain about.

\begin{figure*}[t!]
     \centering
     \includegraphics[width=0.99\textwidth]{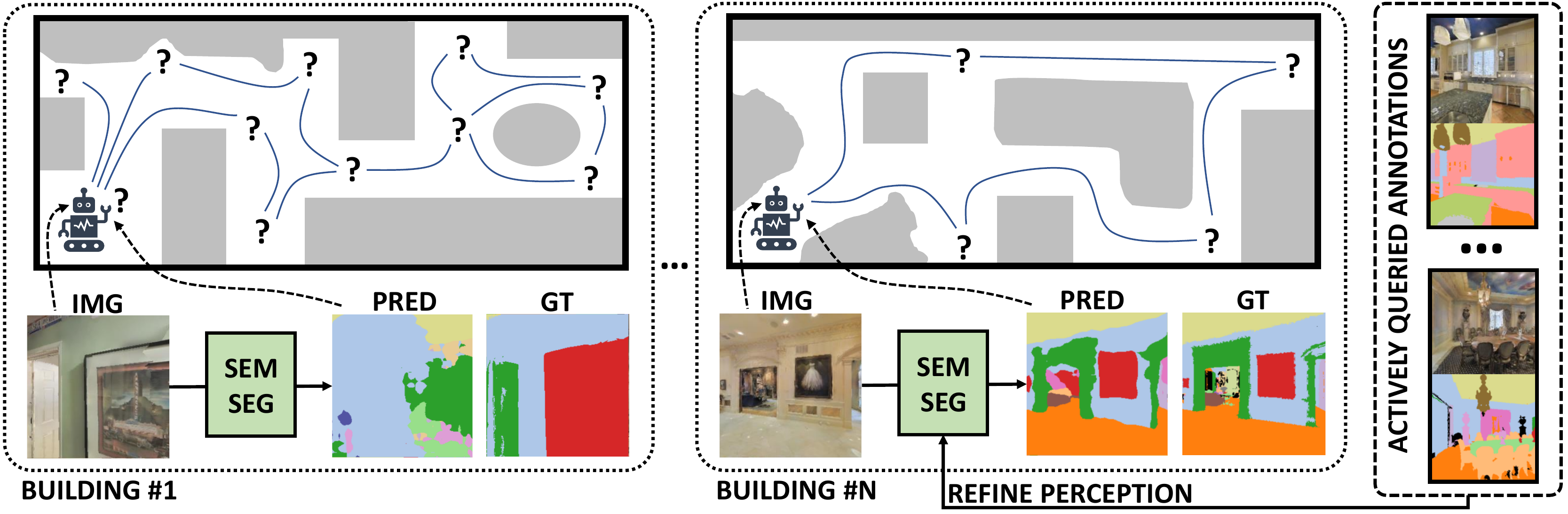}
     \caption{Overview of the embodied lifelong learning setup. A first-person agent selectively determines viewpoints for which to query annotations as it actively explores a new building. Its visual perception system -- here a semantic segmentation network -- is continually refined as the agent queries for more data. As the agent operates in a lifelong framework, its exploration and active learning may differ in later buildings compared to earlier ones. For example, the number of annotations (indicated in the respective maps with question marks), as well as the exploration trajectories, may vary. In this example, in the first building (left), the perception of the agent's current view is poor, and the agent decides to query for the ground truth to refine its perception before continuing to explore. Comparing with a later building (right), the agent's corresponding viewpoint is much more accurate -- potentially because it already learnt to recognize e.g. the 'painting' category (red in the ground truth masks) in the first building -- and in this case the agent decides to ignore querying for that annotation.}
     \label{fig:task-overview}
\end{figure*}

In this paper we thoroughly study this proposed \emph{embodied lifelong learning} task. We present, evaluate and compare agents of various degrees of sophistication -- providing both heuristic alternatives and a fully learnable one -- and with varying amounts of prior visual experience. An overview of the task is shown in \Figure{fig:task-overview}. We let visual perception be measured as semantic segmentation accuracy, given that semantic segmentation is a core perception task that requires fine (pixel-wise) predictions of each view, but note that our framework is applicable in the context of any perception task. To this end, we equip an agent with a semantic segmentation network and the agent is then tasked to move around and query for annotations in order to refine it. After exploring a novel building the agent should be able to accurately segment views in the whole building using a limited number of annotations. Note that which and how many viewpoints should be annotated may be affected by the agent's prior visual perception performance obtained from earlier explored buildings. To empirically investigate this new task, we compare and evaluate all the proposed methods in the Matterport3D simulator \cite{Matterport3D} which provides photorealistic renderings of a large set of indoor scenes. 

In summary, our main contributions are:
\begin{itemize}
    \item We introduce and study the embodied active learning task in a lifelong learning setup where experience and annotations in one scene can guide and inform the exploration and active learning in successive ones.
    \item We introduce several heuristic as well as a fully learnable RL formulation for the task, and study how they are respectively affected by the level of prior visual knowledge.
    \item We provide extensive experimental evaluation of our proposed methods and show that a fully learnable RL agent can use prior visual perception to guide its exploration and active learning more effectively compared to heuristic alternatives, both in terms of exploration granularity and frequency of annotations.
\end{itemize}

\section{Related Work}
Our work is broadly related to embodied learning, active learning, as well as lifelong learning, and we here mention some related work for each subproblem.
\\ \\
{\bf Embodied learning.} With the emergence of large scale simulators \cite{habitat19arxiv,xia2018gibson} of indoor scans \cite{Matterport3D} there has been much recent work on applying RL to visual navigation problems. In point goal navigation \cite{wijmans2019dd} an agent is given a navigation target in polar coordinates $(r,\theta)$ and the task is to reach the goal, which often requires navigating around obstacles. There is a large body of work for this problem, and in general approaches based on RL \cite{wijmans2019dd,sax2020learning} perform well. A related problem is visual exploration where an agent is given egocentric observations and should navigate to explore as much of a scene as possible \cite{chaplot2019learning,ramakrishnan2021exploration}. Another setup related to our paper is studied by \citet{ammirato2017dataset,yang2019embodied} and concerns how to locally explore to get a better view of a partially occluded object. 

How to explore a scene to obtain relevant views for learning visual perception is studied by \citet{chaplot2020semantic}. In their setup the refinement of the visual model is not interactive, but rather a post-processing step -- the visual perception is updated after exploring several scenes, while we refine the visual perception online whenever the agent queries for annotations. Furthermore, in \citet{chaplot2020semantic} all views of the explored trajectories are labelled, while we also task the agent to select only a small subset of frames to label. Different but related  problem setups include learning along fixed navigation trajectories by self-supervision and querying for labels of a subset of the data \cite{song2015robot,pot2018self}, as well as formulations which consider joint improvement of SLAM and visual perception \cite{zhong2018detect,wang2019unified}. Another work related to ours is \citet{nilsson2020embodied}, to which we however differ in several key respects. In our formulation the exploration is global, which is enforced by integrating a global planner with our proposed agents (thus we do not restrict the problem to recognizing only a local neighborhood of where an agent is spawned, as \citet{nilsson2020embodied}). Furthermore, we study the task in a lifelong setup instead of considering independent scenes, and a key contribution of our work is showing how the behavior of agents can improve as they explore multiple scenes consecutively instead of independently and in separation.
\\ \\
{\bf Active learning.} In active learning \cite{settles2009active} the problem is to select which data among a large pool of unlabelled data to label to maximally improve a model. Common approaches include RL \cite{fang2017learning,konyushkova2018discovering}, coverage of a feature space \cite{sener2018active}, uncertainty estimation \cite{gal2017deep,kirsch2019batchbald}, or generative modelling \cite{sinha2019variational,ducoffe2018adversarial}. Our paper differs from the standard active learning formulation in that we do not assume access to a pool of unlabelled data. Instead we consider embodied agents that need to explore an environment to gather the data considered for annotations.

We next briefly review methods for active learning (AL) for semantic segmentation. \citet{siddiqui2020viewal} propose an AL strategy that exploits viewpoint consistency in multi-view datasets. Their method queries labels for those views and superpixels which are not consistently segmented across views from different directions. \citet{golestaneh2020importance} show that self-consistency can greatly improve the performance of a model.
They suggest an AL framework where small image patches that need to be labeled are iteratively extracted by selecting image patches which have high uncertainty (high entropy) under equivariant transformations. The approach by \citet{kasarla2019region} uses entropy- and region-based AL, with annotation at the superpixel level in images, along with the use of fully connected CRFs for label propagation. \citet{mackowiak2018cereals} study region-based AL by estimating labelling cost and uncertainty of unlabelled regions, and \citet{casanova2020reinforced} present a deep reinforcement learning-based approach, where an agent learns a policy to select a subset of image patches to label.
\\ \\
{\bf Lifelong learning.} In typical lifelong learning \cite{parisi2019continual,thrun1995lifelong}, or continual learning, the task is to sequentially train a model for several different tasks in a specific order, while not forgetting the previously learned tasks \cite{kirkpatrick2017overcoming,kemker2018measuring}. Approaches to continual learning for semantic segmentation include storing prototype features of old classes \cite{michieli2021continual}, pooling schemes to preserve spatial relations \cite{douillard2020plop} and strategies to cope with the ambiguity and semantic shift of the background class which in different learning stages can overlap with foreground objects at other stages \cite{cermelli2020modeling}.

\section{Embodied Lifelong Learning}\label{sec:methodology}
We here describe the proposed embodied lifelong learning task in more detail -- see \Figure{fig:task-overview} for an overview. An agent equipped with a visual perception model (here a semantic segmentation network; see \Section{sec:sem-seg}) is spawned in a random location in a building, and its task is to actively move around and explore the environment, while restrictively querying annotations for viewpoints in order to refine its visual perception. To this end, the agent is equipped with four actions \verb|MoveForward|, \verb|RotateLeft|, \verb|RotateRight| and \verb|Annotate|. In \Section{sec:global-nav} we describe how we approach global navigation to ensure systematic exploration of the scenes; we use the same core navigation module within all methods
that we evaluate (see \Section{sec:experiments}). These agents range from heuristic to a fully learnable one. The latter approachs is based on deep RL and is explained in \Section{sec:rl-agent}.

\subsection{Global Navigation}\label{sec:global-nav}
\begin{figure*}
    \centering
    \includegraphics[width=0.99\textwidth]{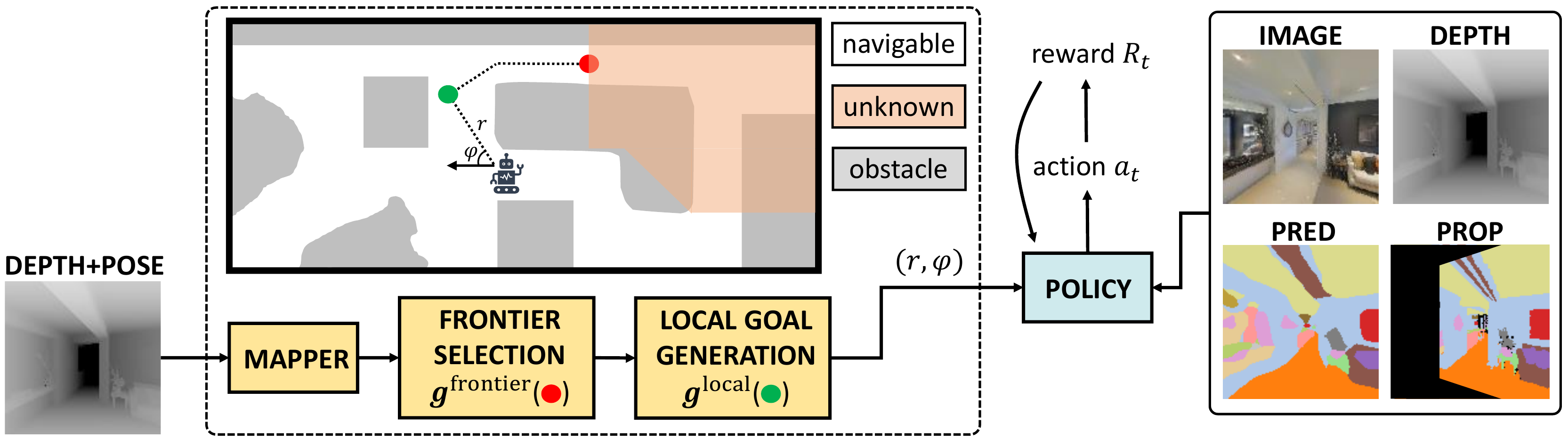}
    \caption{To ensure systematic exploration of a scene we combine a mapper with a local point goal navigation policy. The mapper constructs a 2d top down map of the environment using the agent’s pose and depth, marking each point either as navigable, obstacle or unknown. Each agent (be it heuristic or learnable) uses a hierarchical approach to navigation, where a frontier point $\bs{g}^{\mathrm{frontier}}$ is first selected; then the agent obtains a local goal $\bs{g}^{\mathrm{local}}$ on the shortest path to $\bs{g}^{\mathrm{frontier}}$. For the learnable RL-agent (see \Section{sec:rl-agent}), following prior work on point goal navigation, we give the target location $\bs{g}^{\mathrm{local}}$ as policy input via polar coordinates $(r, \varphi)$, where $r$ and $\varphi$ denote distance and rotation, respectively, to $\bs{g}^{\mathrm{local}}$ relative to the agent’s pose. After each movement the agent is rewarded in proportion to the decrease in geodesic distance to $\bs{g}^{\mathrm{local}}$, as well as by a constant reward once it reaches $\bs{g}^{\mathrm{local}}$, see \eqref{eq:rew-exp}. At this point the mapper provides the agent with a new local navigation target. Note that we also provide rewards related to visual perception refinement, see \eqref{eq:rew-seg}.}
    \label{fig:global_navigation}
\end{figure*}

For an embodied agent to learn a visual perception model that performs well in a given environment, the agent should systematically explore the scene so that it gets an opportunity to observe a diverse set of viewpoints containing various objects. To this end we explicitly map the scene as the agent navigates and provide goal locations for the agent to navigate to. We thus decouple the global navigation from the local exploration and visual learning. A high level overview is given in \Figure{fig:global_navigation}.

The pose and depth at every time step is sent to the mapper which uses the new information to update its map. Using the map, a navigation target is selected to expand the boundary of the map. We use depth and pose to compute a map of the environment. Specifically, given a point $\bs{x}$ in the image plane with depth $d$ we can compute the 3d point $\bs{X}$ using the camera pose. Once we have the 3d points corresponding to the pixels, we threshold based on the height relative to the ground plane to classify as being navigable or an obstacle. For each point on the 2d grid map we keep a counter of the classifications and use majority voting to determine if it is navigable or an obstacle. 

We use classical frontier exploration \cite{yamauchi1998frontier} for the global policy. A frontier point is defined as a navigable point on the map that has a neighbouring point for which it is unknown whether it is navigable or an obstacle. The closest frontier point is always selected, and once reached a new frontier point is selected until no more frontier points remain. To ease the navigation problem, as frontier points might be far away from the current location, we use a local planner as well. Using the map, we compute the shortest path to the selected frontier point $\bs{g}^{\mathrm{frontier}}$ using Dijkstra's algorithm. To select a local navigation target $\bs{g}^{\mathrm{local}}$, we split the shortest path based on curvature, whenever the direction changes, and use this sequence as navigational targets. This was found on the validation set to be more robust than using a fixed distance between local targets.

\subsection{Semantic Segmentation}\label{sec:sem-seg}
For semantic segmentation we use ResNet-50 \cite{he2016deep} pre-trained on ImageNet \cite{JiaDeng2009imagenet}. We make the network fully convolutional \cite{long2015fully} and change the output resolution from $1/32$ to $1/8$ of the input image resolution by removing strides and adding dilations in the ResNet blocks originally having a lower resolution than $1/8$ in the same way as e.g. \citet{chen2017deeplab}. We replace the final classification layer with a new one with the correct number of classes, i.e. 40 in Matterport3D, followed by upsampling by bilinear interpolation to the original input image size.

At the beginning of each episode we either keep the semantic segmentation parameters from the previous episode if we evaluate in the lifelong setup or we initialize all parameters from ImageNet pre-training except for the last classification layer which is initialized randomly. During episodes, we refine the segmentation network whenever the agent selects the \verb|Annotate| action. The image and ground truth that the agent queried for are added to the training set of segmentation network. As in \citet{nilsson2020embodied} we refine the segmentation network either until the accuracy of a minibach is $95 \%$ or for at most $1000$ iterations. The minibatches always include the last image the agent queried for annotation, and we use random left-right flipping and scaling for data augmentation. We train the semantic segmentation network with minibatches of size 4, and we use SGD with learning rate $3 \cdot 10^{-3}$, weight decay $10^{-4}$ and momentum $0.9$.

\begin{table*}
    \centering
    \begin{tabular}{c|c|c|c|c|c}
        \hline Model & Setup & $\Delta A$ / annot & $\Delta A$ / step  & mIoU (1-50) & mIoU (51-100) \\ \hline
        \multirow{2}{*}{Accuracy oracle} & Episodic & 1.162 & 0.056 & 0.306 & 0.429 \\
                                    & Lifelong & 1.256 (+0.094) & 0.056 & 0.340 (+0.034) & 0.448 (+0.019) \\ \hline
        \multirow{2}{*}{RL-agent} & Episodic & 0.876 & 0.057 & 0.286 & 0.397 \\
                                  & Lifelong & 1.070 (+0.194) & 0.059 & 0.324 (+0.038) & 0.407 (+0.010) \\ \hline
        \multirow{2}{*}{Uniform} & Episodic & 1.086 & 0.057 & 0.297 & 0.402 \\
                                 & Lifelong & 1.086 ($\pm 0$) & 0.057 & 0.311 (+0.014) & 0.409 (+0.007) \\ \hline
        \multirow{2}{*}{Random} & Episodic & 1.114 & 0.057 & 0.296 & 0.395 \\
                                & Lifelong & 1.063 (-0.051) & 0.057 & 0.303 (+0.007) & 0.393 (-0.002) \\
        \hline
    \end{tabular}
    \caption{Comparison between episodic and lifelong setups for the accuracy oracle, RL-agent and baselines (uniform, random). We see that the average area explored since the last annotation ($\Delta A /$ annot, $\mathrm{m}^2$) when requesting a new annotation is higher on average in the lifelong compared to the episodic setup for both the accuracy oracle and the RL-agent, indicating adaptive behavior depending on the current segmentation accuracy. This is not the case for the two simpler heuristics. The average navigable area added to the mapper per step ($\Delta A /$ step, $\mathrm{m}^2$) is identical for episodic and lifelong for the accuracy oracle and the two baselines, as is expected since they explore identically, but for the RL-agent we see that in the lifelong setup it on average explores faster. Finally, we see that the improvements of the average mIoU after annotations 1 - 50 and 51 - 100 in the lifelong setup compared to the episodic are larger for the accuracy oracle and RL-agent than for the baselines, and especially for the early part of episodes (annotations 1 - 50). This indicates that both the RL-agent and the accuracy oracle adapt their annotation strategies when evaluated in the lifelong setup.}
    \label{tab:ep_vs_ll}
\end{table*}

\subsection{Reinforcement Learning for Embodied Lifelong Learning}\label{sec:rl-agent}
In this section we describe our proposed reinforcement learning (RL) agent for the embodied lifelong learning task. The global navigation procedure in \Section{sec:global-nav} is set as a prior for the RL-agent's exploration. We next describe in detail the policy of the RL-agent and then explain the reward functions used during training.
\\ 
\noindent\textbf{Policy overview.} To integrate local navigation with RL we follow prior work on point goal navigation \cite{habitat19arxiv}. To this end, we specify the current local navigation target $\bs{g}^\mathrm{local}_t$ by $(r_t, \varphi_t)$ which is the position in polar coordinates of $\bs{g}^{\mathrm{local}}_t$ relative to the agent's pose. We add $(r_t, \varphi_t)$ to the state space and use it as input to the deep stochastic policy $\pith(a_t|s_t)$, where we recall that the actions are \verb|MoveForward|, \verb|RotateLeft|, \verb|RotateRight| and \verb|Annotate|. The state space $s_t$ further contains images and segmentations. Specifically, we let $s_t=\{\bs{I}_t, \bs{S}_t, \bs{P}_t, \bs{D}_t, r_t, \varphi_t\}$ where $\bI_t \in \mathbb{R}^{127 \times 127 \times 3}$ is the agent's current view, $\bs{S}_t=\mathcal{S}_t(\bI_t) \in \mathbb{R}^{127 \times 127 \times 3}$ is the semantic segmentation of $\bI_t$ computed using the current version $\mathcal{S}_t$ of the segmentation network, $\bs{P}_t \in \mathbb{R}^{127 \times 127 \times 3}$ is the last annotation propagated to the agent's current pose, and $\bs{D}_t \in \mathbb{R}^{127 \times 127}$ is the depth. Note that we use depth as input to the policy since it is common practice in point goal navigation and often significantly improves the performance \cite{habitat19arxiv,wijmans2019dd}. 
The propagated mask $\bs{P}_t$ is the latest annotation, but propagated to the agent's current view using optical flow. We compute optical flow between consecutive viewpoints using depth and pose. We thus make the same assumptions on available input modalities as the mapper in \Section{sec:global-nav}.
The policy consists of two branches processing the visual information and navigation targets separately, and these are followed by a recurrent network whose hidden state is used to compute the policy. We give further details in the appendix.
\\ \\
\textbf{Rewards.} We here describe the reward function. It consists of two components -- a \emph{perception reward} to encourage sparse annotations of informative viewpoints which improve the overall perception, and an \emph{exploration reward} to encourage the agent to explore the whole building.
\\ 
\underline{Perception reward:} The reward $R^{\mathrm{seg}}_t$ for semantic segmentation is only non-zero when the agent requests annotation ($a_t=\text{Annotate}$), and we define it in that case as
\begin{equation}\label{eq:rew-seg}
\begin{split}
    R^{\mathrm{seg}}_t = \text{mIoU}(\mathcal{S}_{t+1}, \mathcal{R}) - \text{mIoU}(\mathcal{S}_{t}, \mathcal{R}) - \epsilon^{\mathrm{ann}} \\ +\lambda^{\mathrm{acc}}\mathbf{1}(\text{acc}(\mathcal{S}_{t},\bs{I}_t) < \tau^{\mathrm{acc}})
\end{split}
\end{equation}
where $\mathcal{S}_{t+1}$ and $\mathcal{S}_{t}$ are respectively the segmentation networks (i.e. parameter configurations) at time $t+1$ and $t$ (i.e. after and before refinement), $\mathcal{R}$ is the set of reference views in the building, $\epsilon^\mathrm{ann}=0.01$ is a penalty (hyperparameter) paid for each \verb|Annotate| action, $\lambda^\mathrm{acc} = 0.01$ is a hyperparameter, $\mathbf{1}(\cdot)$ evaluates to one if the condition is true and zero otherwise, and $\tau^\mathrm{acc}=0.7$ is a threshold hyperparameter.

\begin{figure*}
    \centering
    \includegraphics[scale=0.55]{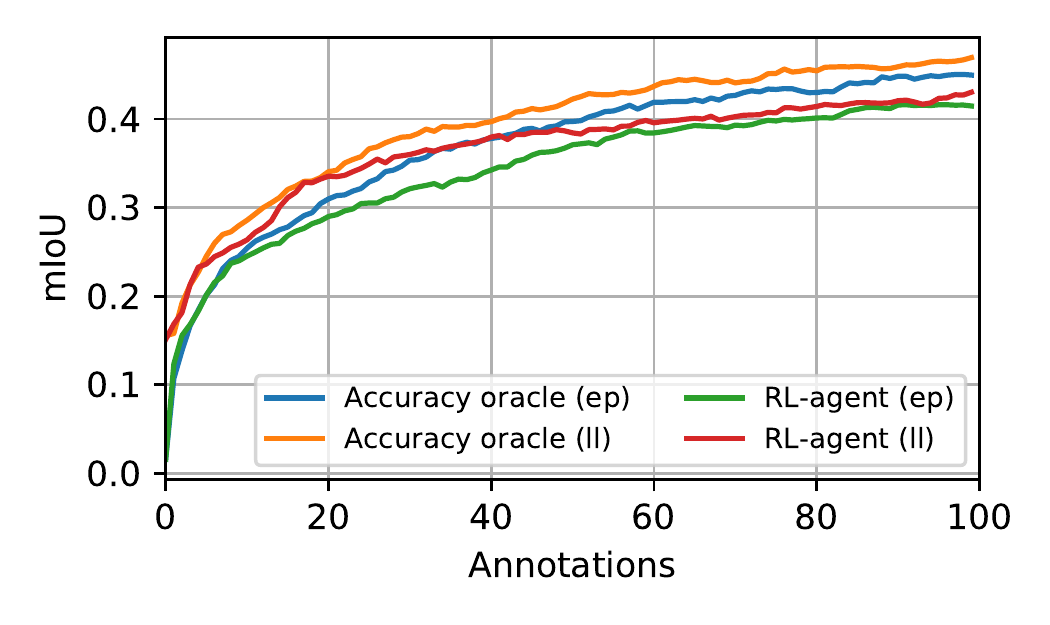}
    \includegraphics[scale=0.55]{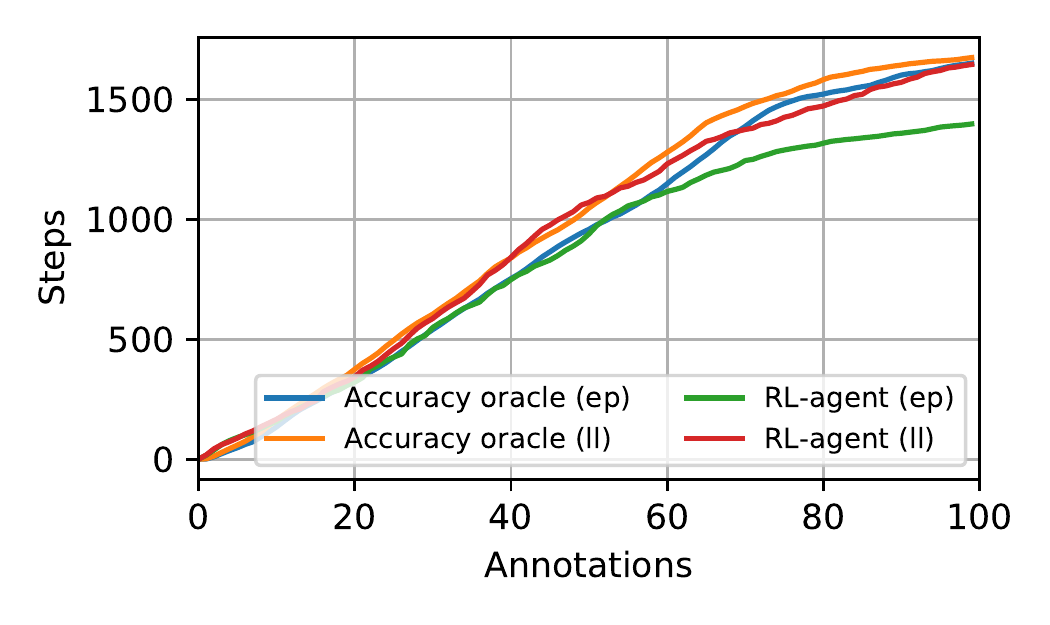}
    \includegraphics[scale=0.55]{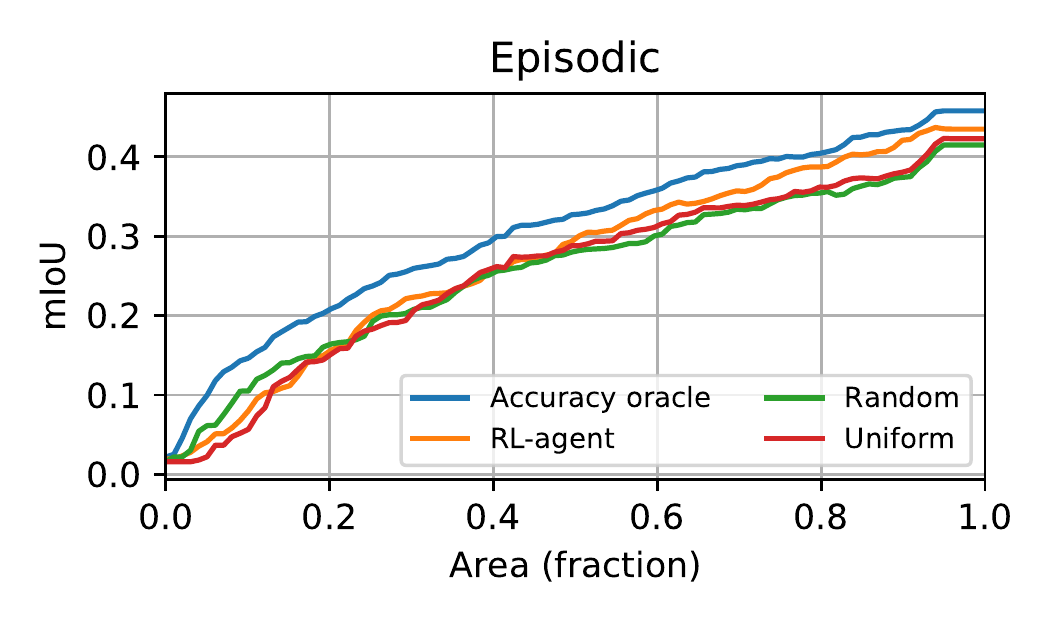} 
    \includegraphics[scale=0.55]{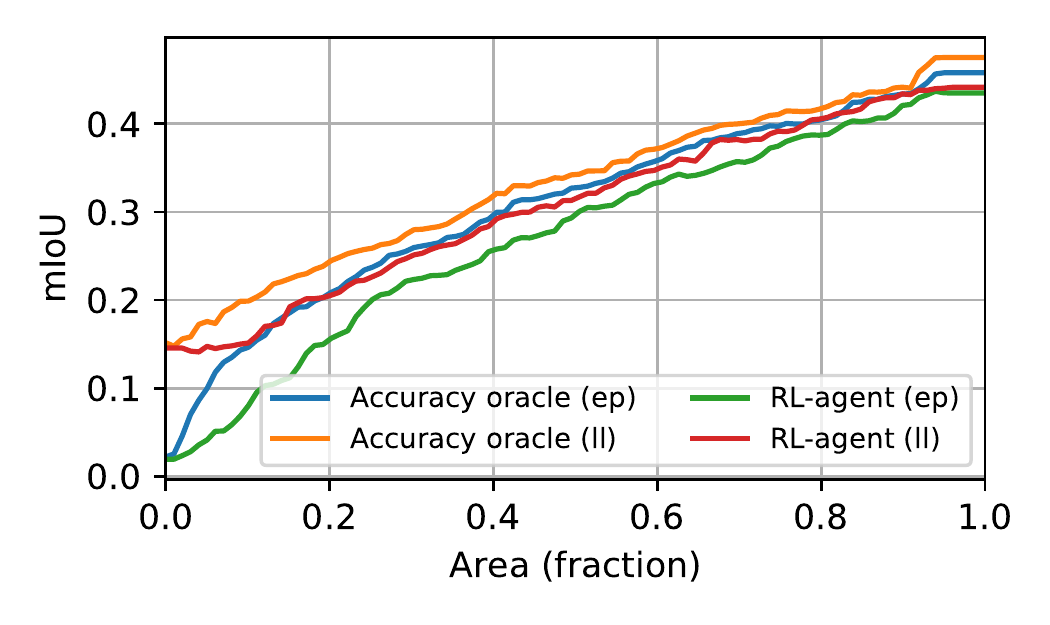}
    \includegraphics[scale=0.55]{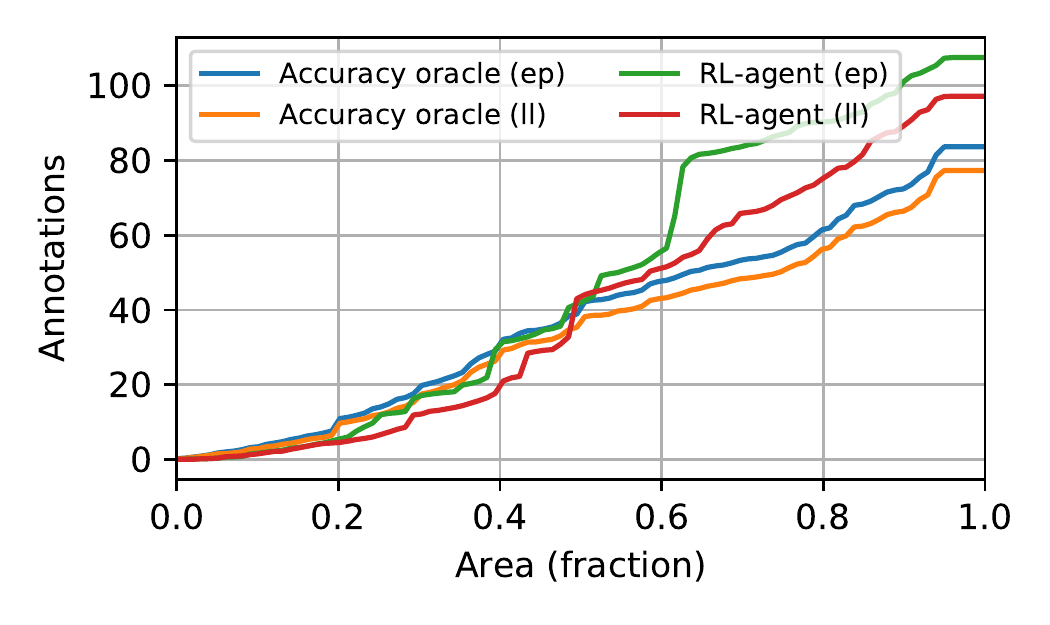}
    \includegraphics[scale=0.55]{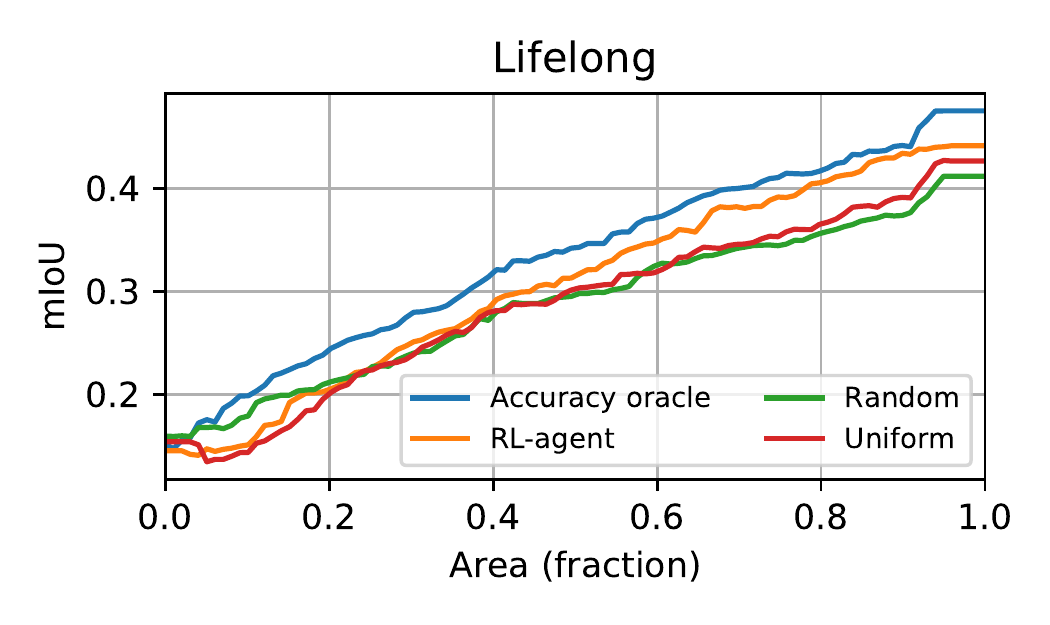} 
    \caption{Comparison between episodic (ep) and lifelong (ll) setups for the RL-agent and the accuracy oracle, and comparison of the RL-agent to the accuracy oracle and heuristic baselines for both the episodic and lifelong setup. In the four left-most plots we see that for a given number of annotations or area explored, both methods get a higher mIoU in the lifelong setup compared to the episodic. We can also see adaptive behavior in the annotation frequencies. For a given number of annotations, both methods take more steps on average  in the lifelong setup than the episodic, indicating less frequent annotations when evaluated in the lifelong setting. We can see the same behavior if we look at the number of annotations for a given explored area -- both methods annotate less frequently in the lifelong setup than in the episodic one. In the two right-most plots we compare all baselines. As expected, the accuracy oracle obtains the highest mIoU. The RL-agent obtains the second highest mIoU at the end of exploration. Moreover, we see a larger improvement of the RL-agent compared to uniform and random in the lifelong setup than in the episodic one.}
    \label{fig:rl_plots}
\end{figure*}

Overall, the perception reward \eqref{eq:rew-seg} is similar to that proposed in \citet{nilsson2020embodied}, but we found it useful for the lifelong setup to include an additional term rewarding asking for annotations when the accuracy $\mathrm{acc}(\mathcal{S}_t, \bs{I}_t)$ of the segmentation network $\mathcal{S}_t$ (prior to refinement) of agent's current view $\bs{I}_t$ was below a threshold. The reference views $\mathcal{R}$ are $32$ randomly sampled views from the entire floor of the scene the agent explores. We compute the mIoU by only selecting the $10$ most common classes in the reference views, to exclude very small objects and unusual classes, since the label distribution is very unbalanced in the Matterport3D scenes.
\\ \\
\underline{Exploration reward:} This reward encourages the agent to move closer to its current local goal $\bs{g}_t^{\mathrm{local}}$ and is given by
\begin{equation}\label{eq:rew-exp}
\begin{split}
    R^{\mathrm{exp}}_t = d(\bx_{t-1}, \bs{g}_t^{\mathrm{local}}) - d(\bx_t, \bs{g}_t^{\mathrm{local}}) \\ +\lambda_g \mathbf{1}(d(\bx_t, \bs{g}_t^{\mathrm{local}}) < \epsilon)
\end{split}
\end{equation}
where $\bx_t$ is the agent's position at step $t$ and $d(\bx, \by)$ is the geodesic distance between the points $\bx$ and $\by$ on the map. We use $\lambda_g = 1$ and $\epsilon = 1$. Note that in contrast to point goal navigation we do not end an episode when a goal is reached, and instead a new goal is given, as described in \Section{sec:global-nav}.
\\ 
\underline{Full reward:} The full reward is given by a convex combination of \eqref{eq:rew-seg} and \eqref{eq:rew-exp}, i.e. $R_t = \lambda R^{\mathrm{exp}}_t + (1 - \lambda) R^{\mathrm{seg}}_t$
with $\lambda = 0.01$. This reward thus balances between visual learning (i.e. improving the agent's perception) and environment exploration.
\\ \\
\textbf{Policy training.} We train the policy using PPO \cite{schulman2017proximal}. To train the policy in the lifelong setup, we reset the segmentation network every $10$th episode, and run $4$ episodes of length $500$ per scene. We pre-train the policy for $10^6$ steps for standard point goal navigation, which does not require refinement of the segmentation model, and we then train the policy in the full segmentation environment for an additional $10^6$ steps which takes about $4$ days using $2$ Nvidia Titan X GPUs. See the appendix for more details and ablations.

\section{Experiments}\label{sec:experiments}
In this section we empirically evaluate various methods for our proposed task. Our main focus is to investigate differences in agent behavior (in terms of exploration and annotations strategies) between the episodic and the lifelong setups, as well as compare how the proposed RL-agent performs compared to several baseline methods for the task. We also provide additional experiments including ablation studies of the RL-agent in the appendix.
\\ \\
\noindent\textbf{Experimental setup.} All methods are evaluated on the Matterport3D dataset \cite{Matterport3D} and we use the simulator framework Habitat \cite{habitat19arxiv} in which an agent can navigate in the photorealistic scenes. We use the official split \cite{Matterport3D} which contains $61$ training scenes, $11$ validation scenes and $18$ test scenes.

When we evaluate on the test set of Matterport3D, we run one episode per scene, for a total of $18$ episodes. In a given scene, all methods we evaluate start at the same position. We end the episodes either when the agent has explored all of the scene or after $2000$ steps. In each scene we train and evaluate the semantic segmentation network  using train and test sets in that scene. The training data is obtained by the queries of the agent we evaluate and the test set is obtained by randomly sampling $32$ views uniformly in the whole scene (we always use the same random views for a given scene when we evaluate different methods). Hence we obtain one mIoU score per scene, and we only use the $10$ most common labels per scene when computing the mIoU due to very unbalanced data. When we report mIoU we always refer to the average over all $18$ test scenes. Similarly, when we report metrics as a function of the exploration, we refer to the area marked as navigable by the mapper (cf. \Section{sec:global-nav}) normalized by the total area (separately within each scene) and then average metrics over all scenes.
\\ \\
{\bf Episodic and lifelong setups.} We differentiate between episodes where the segmentation network is reset at the start of episodes or not by the terms \emph{episodic} and \emph{lifelong}. In the episodic case we reset the segmentation network at the beginning of the episode. Thus no prior segmentation network learning or annotations are kept and the agents begin each episode tabula rasa. In the lifelong setup we keep the parameters of the segmentation network from the previous episode (e.g. at the beginning of the 10th episode the perception system has been trained in nine different previous buildings). Thus the agent will typically be able to recognize some objects and/or semantic classes already at the beginning of the episode, and we expect the agent's exploration to be less granular and that it annotates less frequently since similar scenes or objects might have been seen before. Note that when evaluating the RL-agent (cf. \Section{sec:rl-agent}) in the respective settings, the exact same policy is used to be able to assess the agent's adaptability with respect to its current perception. We use the same ordering of the 18 test scenes for all methods we evaluate, and found that the ordering had a negligible impact (see appendix). 
\\ \\
\textbf{Evaluated methods.} We evaluate the RL-agent (cf. \Section{sec:rl-agent}) and several baselines that make use of the global exploration methodology described in \Section{sec:global-nav}. Since we use a global mapper and planner, we let the baseline agents consist of variants which follow shortest paths to the generated navigation goals. We evaluate such agents with different annotation strategies. The baselines are:
\begin{itemize}
    \item {\bf Accuracy oracle.} Annotates if the semantic segmentation accuracy of its current view is below a threshold (set to $0.7$ based on a validation set). Since this baseline requires access to ground truth segmentations we call it an oracle, and it should be considered an upper bound.
    \item {\bf Uniform annotations.} Annotates with a fixed frequency, which we set to every 20th step to match the average frequency of the accuracy oracle (making them comparable). This is used to show how much can be gained by choice of annotations (as the accuracy oracle) compared to heuristics.
    \item {\bf Random annotations.} Requests annotations randomly with probability 5\% and thus it moves according to the global mapper with 95\% probability. This baseline also annotates with the same average frequency as the accuracy oracle.
\end{itemize}
We further compare to a pre-trained segmentation model using 10,000 random views from the training scenes of Matterport3D. We use 50 of the 61 training scenes for training data and 11 for validation and use the same ResNet model (cf. \Section{sec:sem-seg}) as for all other evaluations.

\subsection{Comparing Lifelong and Episodic Setups} In this section we provide comparisons between the episodic and lifelong setups for the RL-agent and the accuracy oracle, see \Table{tab:ep_vs_ll} and \Figure{fig:rl_plots}. We see that the mIoU is higher at the beginning of episodes, either for the first annotations or for limited observed area, for lifelong compared to episodic. This is expected since in the lifelong setup the semantic segmentation network has already been trained in previously explored scenes, and thus there is some transfer to the current one. Similarly, for both methods we observe differences in annotation frequency between the two settings, e.g. the average area explored per requested annotation is higher in the lifelong setup for both of these methods. We see that while both methods annotate more sparsely in the lifelong setup compared to the episodic, the mIoU is still higher for a comparable number of annotations or level of exploration. %

\begin{figure}
    \centering
    \includegraphics[scale=0.86]{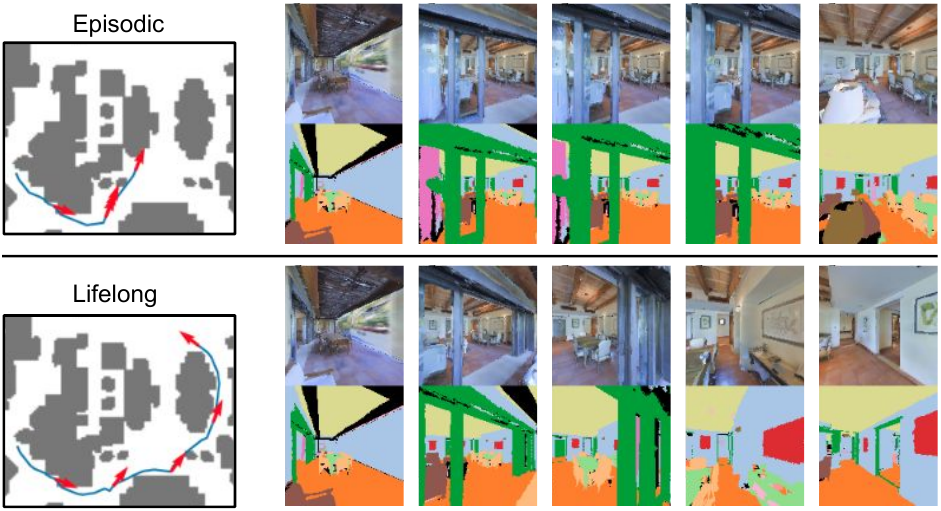}
    \caption{The first five requested annotations when evaluating the RL-agent in the episodic versus lifelong setup. The agent requests annotations at a sparser rate when evaluated lifelong in this scene. For more qualitative examples, please see the appendix.}
    \label{fig:annotations_RL}
\end{figure}

In \Figure{fig:annotations_RL} we show example trajectories of the RL-agent on a scene, and we compare the requested annotations in the episodic and lifelong setups. We see that the RL-agent annotates more sparsely in the lifelong setup than in the episodic one, and thus explores more of the scene per annotation. This can also be seen more generally in \Table{tab:ep_vs_ll} -- the average explored area per annotation is higher in the lifelong setup than the episodic for the RL-agent. Thus the agent is able to adapt its behavior based on its current visual perception performance. 

\subsection{Comparisons with the Heuristic Baselines}\label{sec:exp-with-bls} In \Figure{fig:rl_plots} and \Table{tab:ep_vs_ll} we provide comparisons of the RL-agent to the baselines which explore with frontier exploration and request annotations using different heuristic strategies. By comparing the accuracy oracle, uniform and random, we see that for a given area explored, the accuracy oracle gets a higher mIoU. As seen in \Table{tab:ep_vs_ll}, all methods annotate at a similar frequency with respect to the area, but only for the accuracy oracle do we observe a difference between the area explored per annotation between episodic and lifelong. We also see that the gap in mIoU between the episodic and lifelong setups is greater for the accuracy oracle than uniform and random. These results suggest that uniform and random are unable to adapt with respect to the performance of their perception models. 

If we compare the RL-agent to uniform and random, we see that the mIoU is similar in the beginning of episodes, but at the end of exploration the RL-agent gets a higher mIoU. We also see that there is a larger difference in the lifelong setup. 
The increase in mIoU in the lifelong setup compared to the episodic one is greater for the RL-agent than the heuristic baselines. 
This indicates that the RL-agent is able to adapt its active learning strategy based on its current perception performance. 

\subsection{Pre-training the Segmentation Model}
If we pre-train the segmentation model on the training scenes of Matterport3D we obtain an mIoU of $0.330$ when we evaluate on the test scenes. We note in \Figure{fig:rl_plots} that this performance is comparable to about $20$ annotations in a given scene for the accuracy oracle or RL-agent. The conclusion is that a few ($\sim 20$) scene-specific annotations are as useful as significantly more (10,000) annotations from different buildings. Although surprising, we attribute this to significant appearance differences of the scenes in Matterport3D. This further shows that our proposed task is of importance since a few carefully selected scene-specific annotations are more useful than pre-training with significantly more annotations.

\section{Conclusions}
We have introduced and studied the novel \emph{embodied lifelong learning} task in the context of semantic segmentation, where agents are tasked to explore large buildings and request limited annotations to refine their visual perception. The refinement occurs both within the currently explored building, and improves over the lifetimes of the agents as they tune their perception in an increasing number of buildings. We further introduced a reinforcement learning methodology to jointly learn exploration and visual learning. Our experiments on Matterport3D, covering both the learned method as well as several heuristic ones, show that a trainable method using RL annotates less frequently in the lifelong setup while also achieving more accurate visual perception than when trained from scratch. Overall, different from the compared heuristic alternatives, the learning-based agent exhibits behavior which adapts to the current performance of its visual perception model.

{\small \noindent{\bf Acknowledgments:} This work was supported in part by the European Research Council Consolidator grant
SEED, CNCS-UEFISCDI PN-III-P4-ID-PCE-2016-0535 and PCCF-2016-0180, the
EU Horizon 2020 Grant DE-ENIGMA, Swedish Foundation for Strategic Research (SSF) Smart Systems Program, as well as the Wallenberg AI, Autonomous Systems and Software Program (WASP) funded by the Knut and Alice Wallenberg Foundation.}

\bibliography{aaai22}

\newpage

\appendix

\begin{table*}[h]
    \centering
    \begin{tabular}{c|c|c|c|c|c}
        \hline Model & Setup & $\Delta A$ / annot & $\Delta A$ / step  & mIoU (1-50) & mIoU (51-100) \\ \hline
        \multirow{5}{*}{Accuracy oracle} & Episodic & 1.162 & 0.056 & 0.306 & 0.429 \\
                                    & Lifelong & 1.256 & 0.056 & 0.340 & 0.448 \\ 
                                    & Lifelong, ordering 1 & 1.264 & 0.057 & 0.340 & 0.451 \\ 
                                    & Lifelong, ordering 2 & 1.275 & 0.057 & 0.339 & 0.444 \\ 
                                    & Lifelong, ordering 3 & 1.258 & 0.057 & 0.338 & 0.442 \\ 
        \hline
    \end{tabular}
    \caption{We evaluate the accuracy oracle baseline on three different random scene orderings of the 18 test scenes. The first two rows are identical to that of the results table in the main paper. We see that the metrics are very close in all evaluations in the lifelong setup meaning that the specific ordering used in the main paper has little influence on the results.}
    \label{tab:appendix:scene_orderings}
\end{table*}

\centerline{\LARGE\bf\centering Appendix}

\section{Overview}
In this appendix material we provide additional insights into our proposed embodied lifelong learning task as well as the RL-based approach for modeling the task. Qualitative examples are shown in \Section{sec:appendix:qualitative}, studies of the scene ordering are shown in \Section{sec:appendix:scene_ordering}, ablations of the RL-based agent are performed in \Section{sec:appendix:ablation}, further details of the RL training are given in \Section{sec:appendix:rl_agent} and details of the hyperparameter tuning of the accuracy oracle in \Section{sec:appendix:acc_oracle}.

\section{Qualitative Examples}\label{sec:appendix:qualitative}
\begin{figure*}
    \centering 
    \includegraphics[width=0.99\textwidth]{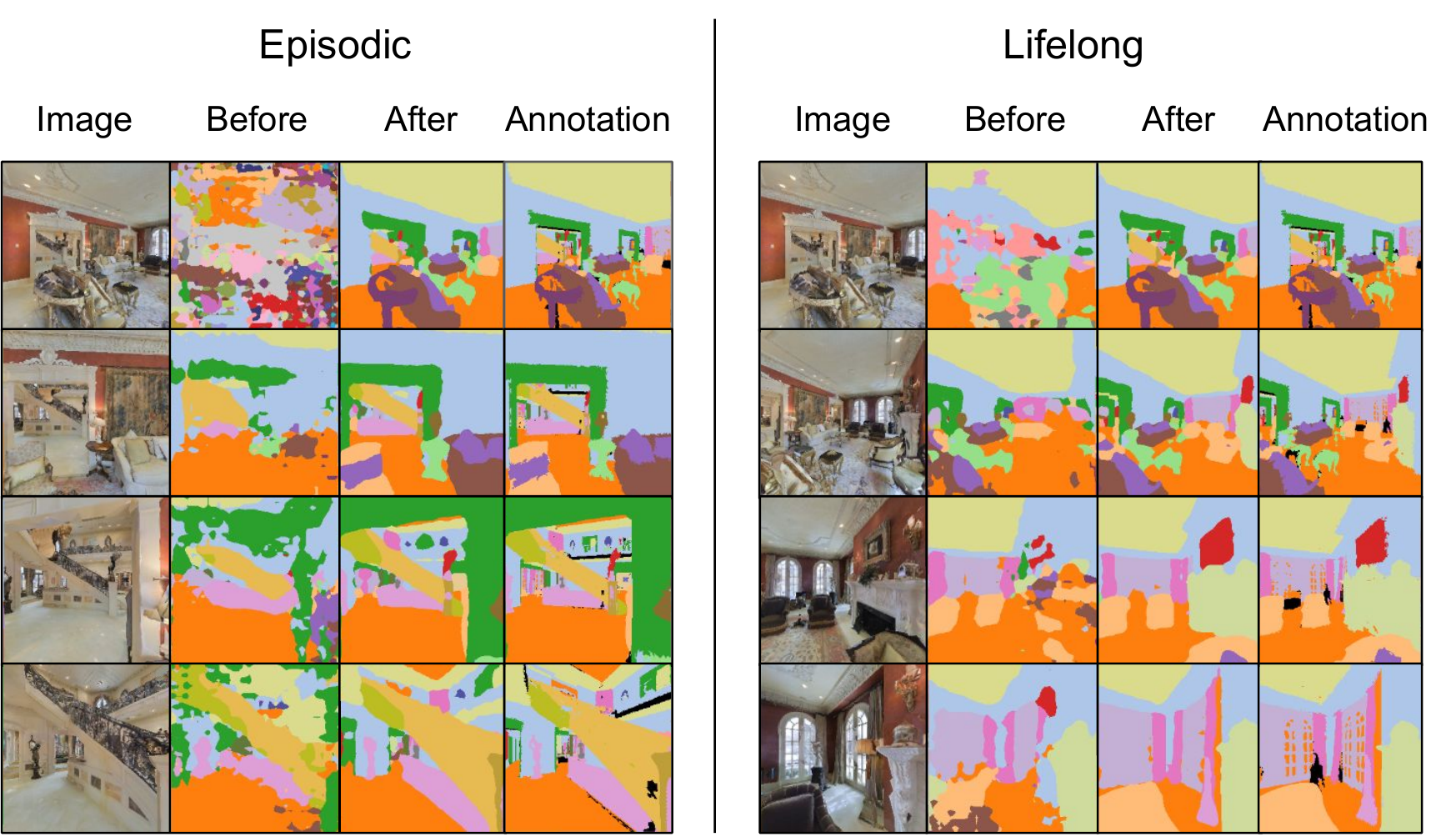}
     \includegraphics[width=0.99\textwidth]{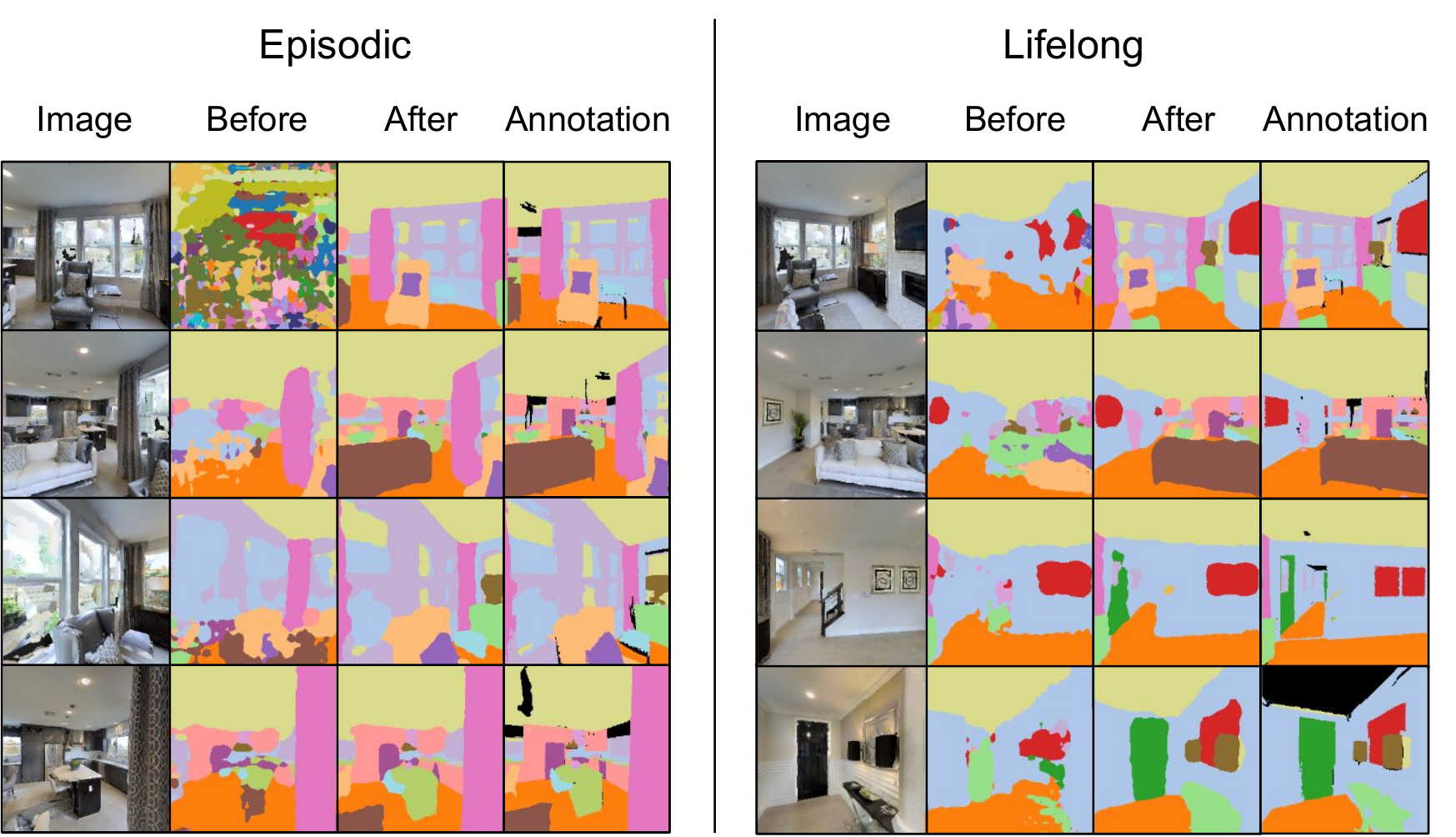}
    \caption{The first four queried annotations, as well as the predicted segmentations before and after refinement, when evaluating the RL-agent in the episodic and lifelong setups in two different buildings from the test set. We can see in the lifelong setup there is some prior knowledge, however it is mostly contained to wall, ceiling and floor. We also note that many of the objects are seen from unusual viewpoints, e.g. the couch in the first images in the first scene. Note e.g. the painting on the left in the second image for the second scene for the lifelong agent. It is correctly recognized despite not being seen in the first annotation.}
    \label{fig:appendix:refinements_RL}
\end{figure*}

We show predicted segmentations before and after refinements in \Figure{fig:appendix:refinements_RL}. As expected, since in the episodic case the classification layer is initialized randomly, the segmentation is inaccurate at the start of episodes, but it is improved after the first annotation. We emphasize that we refine the segmentation model with \emph{one} annotated image at a time, so in the final rows, the segmentation model is trained with just $4$ annotated images. The lifelong agent has prior knowledge, and we can see that some classes e.g. wall, ceiling and floor are fairly accurate in the beginning of episodes, while several smaller objects are not correctly segmented at the beginning but are refined later on, as the segmentation model is retrained with scene-specific annotations that the agent requested.

\begin{figure*}
    \centering 
    \includegraphics[scale=1.72]{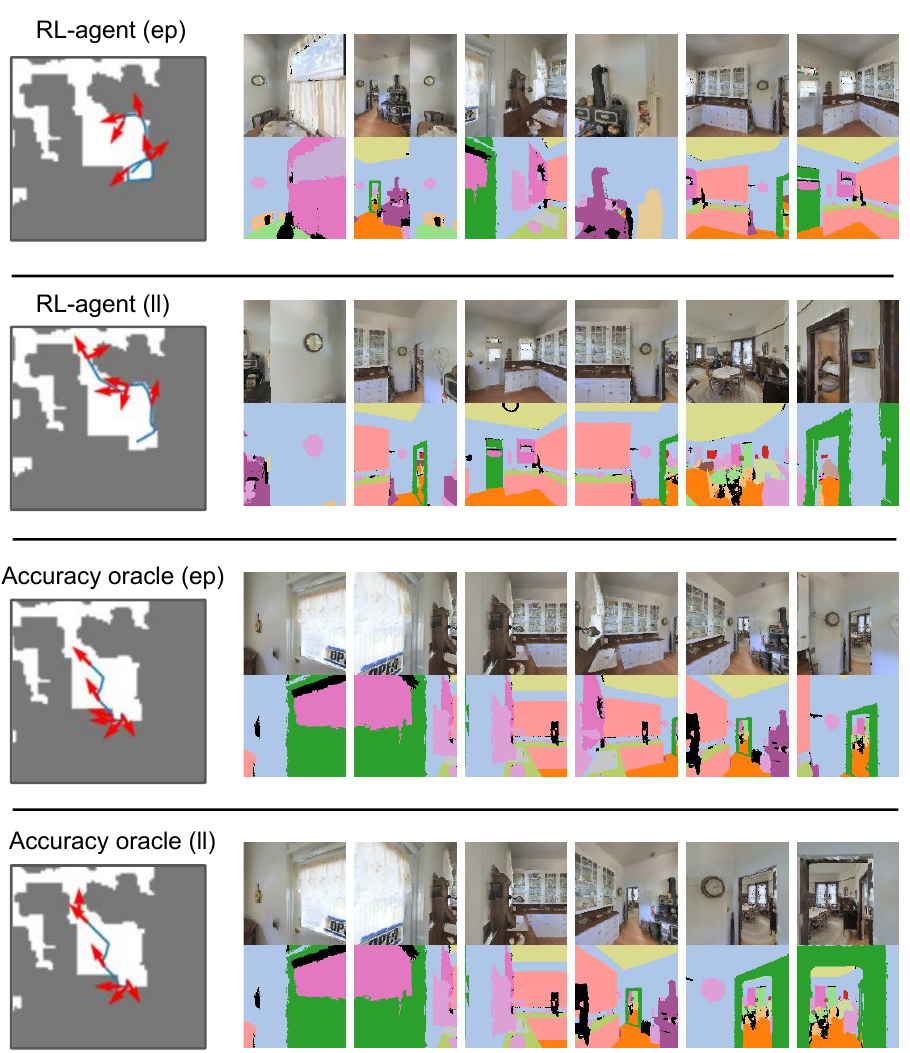}
    \caption{The first six queried annotations when evaluating the accuracy oracle and the RL-agent in the episodic (ep) and lifelong (ll) setups in a building from the test set. We can see that both agents request annotations at a sparser rate when evaluated lifelong in this scene.}
    \label{fig:appendix:annotations_RL}
\end{figure*}

In \Figure{fig:appendix:annotations_RL} we show a qualitative example of how the accuracy oracle and RL-agent annotates in the episodic and lifelong setups. We can see that both methods annotates less frequently in the lifelong setup than the episodic in this scene.

\section{Impact of Scene Orderings}\label{sec:appendix:scene_ordering}

When we evaluate the agents in the lifelong setup we use a fixed ordering of the 18 test scenes which was selected without any special considerations. We use same ordering for all evaluations of different methods in the main paper. To see how robust the performance is if we change the ordering, we try multiple random orderings. In \Table{tab:appendix:scene_orderings} we show results for the accuracy oracle, for episodic and lifelong as in the main paper and additionally for lifelong with $3$ different random orderings of the $18$ scenes. We see that the metrics are very close to the results for lifelong as reported in the main paper, and conclude that the specific ordering has a negligible effect on performance. 

\section{Ablation Studies for the RL-Agent}\label{sec:appendix:ablation}
Various ablations of the RL-agent\footnote{We refer to the RL-agent in the main paper as the \emph{full} RL-agent (or sometimes simply the RL-agent).} proposed in the main paper are given in \Table{tab:appendix:rl_ablation}. We compare to the following versions:
\begin{itemize}
    \item \textbf{Episodic training.} When training the full RL-agent we reset the segmentation network every 10th episode, while for this version we reset it every episode. The agent thus trains in the same way as it is evaluated in the episodic setup.
    \item \textbf{No accuracy reward.} We discard the term giving a reward if the accuracy of the agent's current view is below a threshold. Hence we let $\lambda^{\mathrm{acc}}=0$ in the segmentation reward $R^{\mathrm{seg}}_t$.
    \item \textbf{No navigation pre-training.} We train the policy directly in the full segmentation environment without the navigation pre-training (whereas the full RL-agent was pre-trained for $10^6$ steps for point goal navigation).
    \item \textbf{No global exploration.} We use a spatial coverage reward similar to that in \citet{nilsson2020embodied} instead of the global exploration with point goal navigation to local navigation targets.
\end{itemize}

For consistency with the test results in the main paper we report the average mIoU after 1-50 and 51-100 annotations. However, mIoU(51-100) is slightly inconclusive since the agents sometimes stop before 100 annotations and the annotation frequencies differ significantly, and much more than the methods we compare on the test set in the main paper. Due to this we also report the average number of annotations per episode (rightmost column in \Table{tab:appendix:rl_ablation}). Since more frequent annotations lead to a larger coverage of the objects in the scene we can expect a higher mIoU if the frequency is significantly higher, but this comes at a cost of requiring more annotations. 

We can see that the full RL-agent gets a significantly larger difference in mIoU between the episodic and lifelong setup than the ablated versions. The mIoU is higher in the lifelong setup despite using fewer average number of annotations per episode. Compared to episodic training (resetting the segmentation network every episode) we see that the full RL-agent gets a significantly higher mIoU in the lifelong setup. If we do not use an explicit accuracy reward we see that the annotation frequency is the same in the episodic and lifelong setups, and the difference in mIoU is smaller. If we do not pre-train for navigation the mIoU does not significantly improve in the lifelong setup compared to the episodic. Finally we see that if we use a spatial coverage reward instead of the global navigation with local navigation targets, the exploration is significantly worse, since the average area explored per step ($\Delta A$ / step) is about half of the other methods.

\begin{table*}
    \centering
    \begin{tabular}{c|c|c|c|c|c|c}
        \hline Model & Setup & $\Delta A$ / annot & $\Delta A$ / step  & mIoU (1-50) & mIoU (51-100) & Annots \\ \hline
        \multirow{2}{*}{\makecell{Accuracy \\ oracle}} & Episodic & 1.452 & 0.077 & 0.368 & 0.471 & 65.8 \\
                                    & Lifelong & 1.530 (+0.078) & 0.077 & 0.395 (+0.027) & 0.490 (+0.019) & 62.4 (-3.4) \\ \hline
        \multirow{2}{*}{RL-agent} & Episodic & 0.998 & 0.066 & 0.301 & 0.395 & 95.9 \\
                                    & Lifelong & 1.076 (+0.078) & 0.065 & 0.362 (+0.061) & 0.446 (+0.051) & 84.7 (-11.2) \\ \hline
        \multirow{2}{*}{\makecell{Episodic \\ training}} & Episodic & 0.765 & 0.092 & 0.312 & 0.399 & 131.6 \\
          & Lifelong & 0.856 (+0.091) & 0.071 & 0.336 (+0.024) & 0.424 (+0.025) & 106.2 (-25.4) \\ \hline
         
         \multirow{2}{*}{\makecell{No acc \\ reward}} & Episodic & 0.529 & 0.058 & 0.308 & 0.417 & 143.2 \\
                                  & Lifelong & 0.551 (+0.022) & 0.071 & 0.335 (+0.027) & 0.455 (+0.038) & 147.0 (+3.8) \\ \hline
        \multirow{2}{*}{\makecell{No nav \\ pre-training}} & Episodic & 0.558 & 0.059 & 0.309 & 0.424 & 137.3 \\
                                 & Lifelong & 0.557 (-0.001) & 0.068 & 0.322 (+0.013) & 0.407 (-0.017) & 134.9 (-2.4) \\ \hline
        \multirow{2}{*}{\makecell{No global \\ exploration}} & Episodic & 0.924 & 0.033 & 0.305 & 0.354 & 55.8 \\
                                 & Lifelong & 0.665 (-0.259) & 0.033 & 0.342 (+0.037) & 0.395 (+0.041) & 79.8 (+24.0) \\ \hline
    \end{tabular}
    \caption{Ablation study of the RL-agent proposed in the main paper. These models are evaluated on the 11 validation scenes of Matterport3D. We also include the accuracy oracle for reference. The evaluated metrics are the same as in the main paper for consistency, and we have added a final column (right) which shows the average number of annotations per episode. Note that the full RL-agent (2nd row) queries for significantly fewer annotations per episode in the lifelong compared to the episodic setup, which we can also see for episodic training (3rd row) but not the other ablated versions. Moreover, despite fewer annotations (11.7\% less in lifelong than episodic), the \emph{increase} in mIoU is simultaneously by far the highest for the RL-agent, when looking at lifelong vs episodic.}
    \label{tab:appendix:rl_ablation}
\end{table*}

\section{Further Details of the RL-Agent}\label{sec:appendix:rl_agent}
For the policy training we use a simplified network for semantic segmentation to significantly speed up training. With this modification training is about $5$ times faster than using Resnet-50. We use a nine layer network with three blocks with three convolutional layers each, followed by upsampling and classification. The blocks have 64, 128 and 256 channels respectively, and the first convolutional layer in each block uses stride 2, so the resolutions are $1/2$, $1/4$ and $1/8$ per block. During training we use image size $127 \times 127$ but for all evaluations we use Resnet-50 with image size $256 \times 256$.

The policy consists of two branches processing the visual information and navigation targets separately, and these are followed by a recurrent network whose hidden state is used to compute the policy. The first branch is a four-layer CNN $\phi_{\mathrm{img}}(\bs{I}_t, \bs{S}_t, \bs{P}_t, \bs{D}_t) \in \mathbb{R}^{512}$ which processes the image, segmentations and depth. The second branch is $\phi_{\mathrm{nav}}(r_t, \varphi_t) \in \mathbb{R}^{128}$ which processes the navigation targets. It consists of two fully connected layers of size $64$ and $128$. We concatenate $\phi_{\mathrm{img}}$ and $\phi_{\mathrm{nav}}$ to a vector of length $640$ and use that as an input to an LSTM \cite{hochreiter1997long} with $256$ hidden states. Finally we compute the policy $\pi(a_t|s_t)$ from the hidden state using a fully-connected layer of size $4$ followed by softmax to get a probability distribution over the actions.

The policy network contains $\phi_{\mathrm{img}}$ which processes the image, segmentations and depth. This network consists of 3 convolutional layers, with 32, 64 and 64 channels respectively, filter sizes 8, 4 and 3, and strides 4, 2 and 1. This is followed by global average pooling and a fully-connected layer with 512 units.

We train the policy using PPO \cite{schulman2017proximal} and use the implementation in RLlib \cite{liang2018rllib}. We use learning rate $3 \cdot 10^{-4}$. To train the policy in the lifelong setup, we reset the segmentation network every $10$th episode, and run $4$ episodes of length $500$ per scene. We pre-train the policy for $10^6$ steps for standard point goal navigation, where we replace the predicted segmentation $\bs{S}_t$ and the propagated segmentation $\bs{P}_t$ with the ground truth segmentation in the state space and removed the \verb|Annotate| action. We then train the policy in the full segmentation environment for an additional $10^6$ steps. The pre-training is significantly faster since the segmentation network is not used and not refined. Training $10^6$ steps in the full environment takes about $4$ days using $2$ Nvidia Titan X GPUs, and the pre-training takes 10 hours on a single GPU. 

\section{Accuracy Oracle Thresholding}\label{sec:appendix:acc_oracle}
We empirically found the 0.7 threshold to give the largest mIoU. We show in \Table{tab:appendix:rl_ablation} that the accuracy oracle with the 0.7 threshold obtains mIoU(1-50)=0.395, mIoU(51-100)=0.490 using on average 62.4 annotations on the validation scenes. If we use 0.6 as threshold we get mIoU(1-50)=0.357, mIoU(51-100)=0.426 using on average 47.7 annotations and for 0.8 as threshold we get mIoU(1-50)=0.367, mIoU(51-100)=0.454 using on average 92.0 annotations. With frequent annotations, once an agent reaches 50 annotations it will only have explored a small area, while with infrequent annotations it will annotate sparsely and miss large parts of the scene. 

\end{document}